\documentclass[11pt,letterpaper]{article}
\usepackage{times}
\usepackage{latexsym}
\usepackage{amsfonts,epsfig,graphicx}
\usepackage{amsmath,amssymb,amsthm}
\usepackage{enumerate}
\usepackage[ruled,linesnumbered]{algorithm2e}
\usepackage{amsfonts}
\usepackage{macros}
\usepackage{fullpage}
\usepackage{color}
\usepackage{stmaryrd}
\usepackage{qtree}
\usepackage{tikz}
\usepackage{tikz-qtree}
\usetikzlibrary{positioning}
\usepackage{subcaption}
\usepackage{enumitem}
\setlist{leftmargin=10pt}
\usepackage{emnlp2017}
\emnlpfinalcopy

\makeatletter
\renewcommand{\boxed}[1]{\text{\fboxsep=.2em\fbox{\m@th$\displaystyle#1$}}}
\makeatother

\def\emnlppaperid{982}

\title{Macro Grammars and Holistic Triggering for Efficient Semantic Parsing}

\author{Yuchen Zhang \and Panupong Pasupat \and Percy Liang \\
{\tt \{zhangyuc,ppasupat,pliang\}@cs.stanford.edu}\\
  Computer Science Department, Stanford University}

\begin{document}

\maketitle

\begin{abstract}
To learn a semantic parser from denotations,
a learning algorithm must search over a
combinatorially large space of logical forms
for ones consistent with the annotated denotations.
We propose a new online learning algorithm that searches
faster as training progresses.
The two key ideas are using \emph{macro grammars} to cache the abstract patterns of useful logical forms found thus far,
and \emph{holistic triggering} to efficiently retrieve the most relevant patterns
based on sentence similarity.
On the \wikitables\ dataset, we first expand the search space of
an existing model to improve the state-of-the-art accuracy from 38.7\% to
42.7\%, and then use macro grammars and holistic triggering to achieve an 11x speedup and an accuracy of 43.7\%.
\end{abstract}

\section{Introduction}

We consider the task of learning a semantic parser for question answering
from question-answer pairs~\cite{clarke10world,liang11dcs,berant2013freebase,artzi2013uw,pasupat2015compositional}.
To train such a parser, the learning algorithm must somehow search for \emph{consistent}
logical forms (i.e.,~logical forms that execute to the correct answer
denotation).
Typically, the search space is defined
by a compositional grammar over logical forms
(e.g.,~a context-free grammar),
which we will refer to as the \emph{base grammar}.

To cover logical forms that answer complex questions, the base grammar must be quite general and compositional,
leading to a huge search space that contains many useless logical forms.
For example, the parser of \citet{pasupat2015compositional} on
Wikipedia table questions (with beam size 100) generates and featurizes
an average of 8,400 partial logical forms per example.
Searching for consistent logical forms is thus a major computational bottleneck.

\begin{table}[t]
\centering\small
\begin{tabular}{@{\hspace*{-1.2em}}c|c|c|c|c|c|}
\cline{2-6}
& \textbf{Rank} &	\textbf{Nation}	& \textbf{Gold}	& \textbf{Silver}	& \textbf{Bronze} \\\cline{2-6}
$r_1:$& 1	&France	&3	&1	&1	\\\cline{2-6}
$r_2:$& 2	&Ukraine	&2	&1	&2	\\\cline{2-6}
$r_3:$& 3	&Turkey	&2	&0	&1	\\\cline{2-6}
$r_4:$& 4	&Sweden	&2	&0	&0	\\\cline{2-6}
$r_5:$& 5	&Iran	&1	&2	&1	\\\cline{2-6}
\end{tabular}
\caption{A knowledge base for the question $x$ = ``\emph{Who ranked right after Turkey?}''. The target denotation is $y$ = \ttt{\{Sweden\}}.}
\label{table:wikitables-example}
\end{table}

In this paper, we propose
\emph{macro grammars} to bias the search towards
structurally sensible logical forms.
To illustrate the key idea, suppose we managed to parse the
utterance ``\emph{Who ranked right after Turkey?''}
in the context of Table~\ref{table:wikitables-example}
into the following consistent logical form (in lambda DCS)
(Section~\ref{sec:kb-and-lf}):
\begin{align*}
\tt {\reverse{Nation}.\reverse{Next}.Nation.Turkey},
\end{align*}
which identifies the cell under the Nation column in the row after Turkey.
From this logical form, we can abstract out all relations and entities to produce
the following \emph{macro}:
\begin{align*}
\reverse{\{Rel\#1\}}.\reverse{\tt Next}.\{Rel\#1\}.\{Ent\#2\},
\end{align*}
which represents the abstract computation: ``identify the cell under the $\{Rel\#1\}$ column in the row after $\{Ent\#2\}$.''
More generally, macros capture the overall shape of computations in a way that
generalizes across different utterances and knowledge bases.
Given the consistent logical forms of utterances parsed so far,
we extract a set of macro rules.
The resulting macro grammar consisting of these rules generates only logical
forms conforming to these macros, which is a much smaller and higher precision
set compared to the base grammar.

Though the space of logical forms defined by the macro grammar is smaller,
it is still expensive to parse with them
as the number of macro rules grows with the number of training examples.
To address this, we introduce \emph{holistic triggering}:
for a new utterance, we find the $K$ most similar utterances
and only use the macro rules induced from any of their consistent logical forms.
Parsing now becomes efficient as only a small subset of macro rules are triggered for any utterance.
Holistic triggering can be contrasted with the norm in semantic parsing,
in which logical forms are either triggered by specific phrases (anchored) or
can be triggered in any context (floating).

Based on the two ideas above, we propose an online algorithm for jointly inducing a macro grammar
and learning the parameters of a semantic parser.
For each training example, the algorithm first attempts
to find consistent logical forms using holistic triggering on the current macro
grammar. If it succeeds, the algorithm uses the consistent logical forms found to
update model parameters. Otherwise, it applies the base grammar for a more
exhaustive search to enrich the macro grammar.
At test time, we only use the learned macro grammar.

We evaluate our approach on the \wikitables~dataset
\cite{pasupat2015compositional}, which features a semantic parsing task
with open-domain knowledge bases and complex questions. We first extend the
model in \newcite{pasupat2015compositional} to achieve a new state-of-the-art
test accuracy of 42.7\%, representing a 10\% relative improvement over the best
reported result~\cite{haug2017neural}.
We then show that training with macro grammars yields an 11x speedup
compared to training with only the base grammar. At test time, using the learned macro grammar achieves
a slightly better accuracy of 43.7\% with a 16x run time speedup over using the base
grammar.

\section{Background}
\label{sec:background}

We base our exposition on the task of question answering on a knowledge base.
Given a natural language utterance $x$, a semantic parser maps the utterance
to a logical form $z$. The logical form is executed on a knowledge base
$w$ to produce denotation $\db{z}_w$. The goal is to train a semantic parser from a training set of utterance-denotation pairs.

\subsection{Knowledge base and logical forms}
\label{sec:kb-and-lf}

A knowledge base refers to a collection of entities and relations. For the
running example ``\emph{Who ranked right after Turkey?}'', we use
Table~\ref{table:wikitables-example} from Wikipedia as the knowledge
base. Table cells (e.g.,~\ttt{Turkey}) and rows (e.g.,~$r_3=$ the 3rd row) are
treated as entities. Relations connect entities: for
example, the relation \ttt{Nation} maps $r_3$ to \ttt{Turkey}, and a special
relation \ttt{Next} maps $r_3$ to $r_4$.

A logical form $z$ is a small program that can be executed on the knowledge base.
We use lambda DCS~\cite{liang2013lambda} as the language of logical forms. The smallest
units of lambda DCS are entities (e.g.,~\ttt{Turkey}) and
relations (e.g.,~\ttt{Nation}). Larger logical forms
are composed from smaller ones, and
the denotation of the new logical form can be computed
from denotations of its constituents. For example,
applying the join operation on \ttt{Nation}
and \ttt{Turkey} gives \ttt{Nation.Turkey}, whose denotation
is $\db{\tt{Nation.Turkey}}_w = \{r_3\}$, which corresponds
to the 3rd row of the table. The
partial logical form \ttt{Nation.Turkey} can then be used
to construct a larger logical form:
\begin{align}\label{eqn:logical-form-example}
 z = \tt {\reverse{Nation}.\reverse{Next}.Nation.Turkey},
\end{align}
where $\reverse{\cdot}$ represents the reverse of a relation. The denotation of the logical form $z$ with respect to the knowledge base  $w$ is equal to $\db{z}_w = \{\tt{Sweden}\}$. See \citet{liang2013lambda} for more details about the semantics of lambda DCS.

\subsection{Grammar rules}
\label{sec:generic-rules}

\newcommand{\NN}[2]{#1\\{\scriptsize #2}}
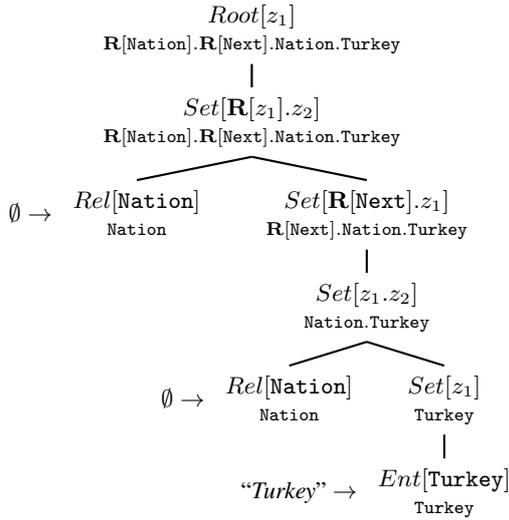
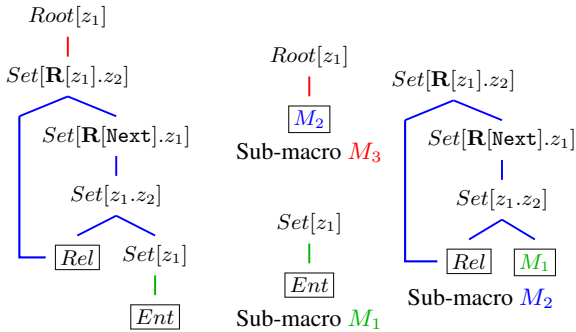
\begin{figure}[t!]
\begin{subfigure}[b]{1.0\linewidth}\centering
\begin{tikzpicture}[
  font=\small,
  every tree node/.style={align=center,anchor=north},
  edge from parent/.append style={thick},
  level distance=35pt,
]
\Tree [.\NN{$Root[z_1]$}{\scriptsize $\tt \reverse{Nation}.\reverse{Next}.Nation.Turkey$}
        [.\NN{$Set[\reverse{z_1}.z_2]$}{$\tt \reverse{Nation}.\reverse{Next}.Nation.Turkey$}
          \node(nation1){\NN{$Rel[{\tt Nation}]$}{$\tt Nation$}};
          [.\NN{$Set[{\tt \reverse{Next}}.z_1]$}{$\tt \reverse{Next}.Nation.Turkey$}
            [.\NN{$Set[z_1.z_2]$}{$\tt Nation.Turkey$}
              \node(nation2){\NN{$Rel[{\tt Nation}]$}{$\tt Nation$}};
              [.\NN{$Set[z_1]$}{$\tt Turkey$}
                \node(turkey){\NN{$Ent[{\tt Turkey}]$}{$\tt Turkey$}}; ]]]]]
\node[left=0 of nation1,anchor=east] {$\emptyset \to$};
\node[left=0 of nation2,anchor=east] {$\emptyset \to$};
\node[left=0 of turkey,anchor=east] {``\emph{Turkey}'' $\to$};
\end{tikzpicture}
\caption{Derivation tree ($z_i$ represents the $i$th child)}
\label{fig:trees-a}
\end{subfigure}
\\[.5em]
\begin{subfigure}[b]{0.33\linewidth}\centering
\begin{tikzpicture}[
  scale=0.9,
  font=\small,
  every tree node/.style={align=center,anchor=north},
  edge from parent/.append style={thick},
  level distance=25pt,
]
\Tree [.$Root[z_1]$
        \edge[red];
        [.$Set[\reverse{z_1}.z_2]$
          \edge[blue];
          \node(n1){};
          \edge[blue];
          [.$Set[{\tt \reverse{Next}}.z_1]$
            \edge[blue];
            [.$Set[z_1.z_2]$
              \edge[blue];
              \node(n2){$\boxed{Rel}$};
              \edge[blue];
              [.$Set[z_1]$
                \edge[green!70!black];
                $\boxed{Ent}$ ]]]]]
\draw[thick,blue] (n2.west) -| (n1.north);
\end{tikzpicture}
\caption{Macro}
\label{fig:trees-b}
\end{subfigure}
~
\begin{subfigure}[b]{0.68\linewidth}\centering
\begin{tikzpicture}[
  scale=0.9,
  font=\small,
  every tree node/.style={align=center,anchor=north},
  edge from parent/.append style={thick},
  level distance=25pt,
]
\begin{scope}[yshift=0]
\Tree [ \edge[red]; $\boxed{\color{blue} M_2}$ ].$Root[z_1]$
\node at (0,-1.7) {\footnotesize Sub-macro {\color{red} $M_3$}};
\end{scope}
\begin{scope}[yshift=-70]
\Tree [ \edge[green!70!black]; $\boxed{Ent}$ ].$Set[z_1]$
\node at (0,-1.7) {\footnotesize Sub-macro {\color{green!70!black} $M_1$}};
\end{scope}
\begin{scope}[xshift=60,yshift=-10]
\Tree [.$Set[\reverse{z_1}.z_2]$
        \edge[blue];
        \node(n1){};
        \edge[blue];
        [.$Set[{\tt \reverse{Next}}.z_1]$
          \edge[blue];
          [.$Set[z_1.z_2]$
            \edge[blue];
            \node(n2){$\boxed{Rel}$};
            \edge[blue];
            $\boxed{\color{green!70!black} M_1}$ ]]]
\draw[thick,blue] (n2.west) -| (n1.north);
\node at (0.4,-3.5) {\footnotesize Sub-macro {\color{blue} $M_2$}};
\end{scope}
\end{tikzpicture}
\caption{Atomic sub-macros}
\label{fig:trees-c}
\end{subfigure}
\caption{From the derivation tree (a),
we extract a macro (b),
which can be further decomposed into atomic sub-macros (c).
Each sub-macro is converted into a macro rule.
}
\label{fig:trees}
\end{figure}

The space of logical forms is defined recursively by grammar rules. In this
setting, each constructed
logical form belongs to a category (e.g., $Entity$, $Rel$, $Set$), with a
special category $Root$ for complete logical forms. A rule specifies the
categories of the arguments, category of the resulting logical form, and how
the logical form is constructed from the arguments.
For instance, the rule
\begin{align}\label{eqn:rule-example}
Rel[z_1] + Set[z_2] \to Set[z_1.z_2]
\end{align}
specifies that a partial logical form $z_1$ of category $Rel$ and $z_2$ of
category $Set$ can be combined into $z_1.z_2$ of category $Set$.
With this rule, we can construct \ttt{Nation.Turkey} if we have constructed
\ttt{Nation} of type $Rel$ and \ttt{Turkey} of type $Set$.

We consider the rules used by~\citet{pasupat2015compositional} for their
floating parser.\footnote{Their grammar and our implementation use more
fine-grained categories ($Atomic$, $Values$, $Records$) instead of $Set$. We
use the coarser category here for simplicity.} The rules are divided into
\emph{compositional rules} and \emph{terminal rules}.
Rule~(\ref{eqn:rule-example}) above is an example of a compositional rule,
which combines one or more partial logical forms together.
A terminal rule has one of the following forms:
\begin{align}
  TokenSpan[span] &\to c[f(span)] \label{eqn:terminal-anchored}\\
  \emptyset &\to c[f(\emptyset)] \label{eqn:terminal-floating}
\end{align}
where $c$ is a category. A rule with the form (\ref{eqn:terminal-anchored})
converts an utterance token span (e.g., ``\emph{Turkey}'') into a partial
logical form (e.g., $\tt{Turkey}$). A rule with the form
(\ref{eqn:terminal-floating}) generates a partial logical form without any
trigger. This allows us to generate logical predicates that do not correspond to any
part of the utterance (e.g., $\tt{Nation}$).

A complete logical form is generated by recursively applying rules. We can
represent the derivation process by a \emph{derivation tree}
such as in
Figure~\ref{fig:trees-a}. Every node of the derivation tree corresponds to one
rule. The leaf nodes correspond to terminal rules, and the intermediate nodes
correspond to compositional rules.

\subsection{Learning a semantic parser}
\label{sec:beam-search-and-learning}

Parameters of the semantic parser are learned from training data
$\{(x_i,w_i,y_i)\}_{i=1}^n$. Given a training example with an utterance $x$, a
knowledge base $w$, and a target denotation $y$, the learning algorithm
constructs a set of candidate logical forms indicated by $\Z$. It then
extracts a feature vector $\phi(x,w,z)$ for each $z\in \Z$, and defines a
log-linear distribution over the candidates $z$:
\begin{align}\label{eqn:scoring-function}
	p_\theta(z \mid x,w)\propto \exp(\theta^\top \phi(x,w,z)),
\end{align}
where $\theta$ is a parameter vector. The straightforward way to construct $\Z$
is to enumerate all possible logical forms induced by the grammar. When the
search space is prohibitively large, it is a common practice to use beam
search. More precisely, the algorithm constructs partial logical forms
recursively by the rules, but for each category and each search depth, it keeps
only the $B$ highest-scoring logical forms according to the model
probability~\eqref{eqn:scoring-function}.

During training, the parameter $\theta$ is learned by maximizing the
regularized log-likelihood of the correct denotations:
\begin{align}\label{eqn:objective-function}
	J(\theta) = \frac{1}{n}\sum_{i=1}^n \log p_\theta(y_i\mid x_i,w_i) - \lambda \lones{\theta},
\end{align}
where the probability $p_\theta(y_i\mid x_i,w_i)$ marginalizes over the space of candidate logical forms:
\[
	p_\theta(y_i\mid x_i,w_i) = \sum_{z\in \Z_i: \db{z}_{w_i} = y_i} p_\theta(z\mid x_i,w_i).
\]
The objective is optimized using AdaGrad~\cite{duchi10adagrad}. At test time,
the algorithm selects a logical form $z\in \Z$ with the highest model
probability~\eqref{eqn:scoring-function}, and then executes it on the knowledge
base $w$ to predict the denotation $\db{z}_w$.

\section{Learning a macro grammar}
\label{sec:learning}

\begin{algorithm}[t]
\KwData{example $(x,w,y)$, macro grammar, base grammar with terminal rules $\mathcal{T}$}
Select a set $\mathcal{R}$ of macro rules\label{step:trigger} (Section~\ref{sec:trigger-macro-rules})\;
Generate a set $\Z$ of candidate logical forms from rules $\mathcal{R}\cup\mathcal{T}$ \label{step:generate} (Section~\ref{sec:beam-search-and-learning})\;
\eIf{$\Z$ contains consistent logical forms}{
Update model parameters (Section~\ref{sec:train-and-predict});
}
{Apply the base grammar to search for a consistent logical form (Section~\ref{sec:beam-search-and-learning})\label{step:apply-base-grammar}\;
Augment the macro grammar (Section~\ref{sec:update-macro-grammar});}
Associate utterance $x$ with the highest- scoring consistent logical form found\; \label{step:associate}
\caption{Processing a training example}\label{alg:learn-macro-grammar}
\end{algorithm}

The base grammar usually defines a large search space containing many irrelevant logical forms.
For example,
the grammar in \citet{pasupat2015compositional} can generate long chains of join operations
(e.g., $\tt \reverse{Silver}.Rank.\reverse{Gold}.Bronze.2$)
that rarely express meaningful computations.

The main contribution of this paper is a new algorithm to speed up the search
based on previous searches.
At a high-level,
we incrementally build a \emph{macro grammar}
which encodes useful logical form macros discovered during training.
Algorithm~\ref{alg:learn-macro-grammar} describes how our learning algorithm
processes each training example.
It first tries to use an appropriate subset of rules in the macro grammar
to search for logical forms.
If the search succeeds, then the semantic parser parameters are updated as usual.
Otherwise, it falls back to the base grammar,
and then add new rules to the macro grammar based on the consistent logical form found.
Only the macro grammar is used at test time.

We first describe macro rules and how they are generated
from a consistent logical form.
Then we explain the steps of the training algorithm in detail.

\subsection{Logical form macros}
\label{sec:derive-macro}

A \emph{macro} characterizes an abstract logical form structure.
We define the macro for any given logical form $z$ by transforming its
derivation tree as illustrated in Figure~\ref{fig:trees-b}. First, for each terminal rule
(leaf node), we substitute the rule by a placeholder, and name it with
the category on the right-hand side of the rule. Then we merge leaf nodes
that represent the same partial logical form. For example, the logical
form~\eqref{eqn:logical-form-example} uses the relation \ttt{Nation} twice, so
in Figure~\ref{fig:trees-b}, we merge the two leaf nodes to impose such a
constraint.

While the resulting macro
may not be tree-like, we call each node \emph{root} or \emph{leaf} if it is a
root node or a leaf node of the associated derivation tree.

\subsection{Constructing macro rules from macros}
\label{sec:derive-rules}

For any given macro $M$, we can construct a set of \emph{macro rules} that,
when combined with terminal rules from the base grammar,
generates exactly the logical forms that satisfy the macro $M$. The
straightforward approach is to associate a unique rule with each macro:
assuming that its $k$ leaf nodes contain categories
$c_1,\dots,c_k$, we can define a rule:
\begin{align}\label{eqn:atomic-macro-rule}
\hspace{-5pt}c_1[z_1] + \dots + c_k[z_k] \to Root[f(z_1,\dots,z_k)],
\end{align}
where $f$ substitutes $z_1,\dots,z_k$ into the corresponding leaf nodes of macro $M$.
For example, the rule for
the macro in Figure~\ref{fig:trees-b} is
\[
	Rel[z_1] + Ent[z_2] \to Root[\reverse{z_1}.\reverse{\tt Next}.z_1.z_2].
\]

\subsection{Decomposed macro rules}
\label{sec:decompose-macro-rules}

Defining a unique rule for each macro is computationally suboptimal
since the common structures shared among macros are not being exploited.
For example, while
$\tt max(\reverse{Rank}.Gold.Num.2)$
and
$\tt \reverse{Nation}.argmin(Gold.Num.2,\ Index)$
belong to different macros, the partial logical form $\tt Gold.Num.2$ is
shared, and we wish to avoid generating and featurizing it more than once.

In order to reuse such shared parts, we decompose macros
into \emph{sub-macros} and define rules based on them. A subgraph $M'$ of $M$ is a sub-macro
if (1) $M'$ contains at least one non-leaf node;
and (2) $M'$ connects to the rest of the macro $M\backslash M'$ only through one node (the root of $M'$).
A macro $M$ is called \emph{atomic} if the only sub-macro of $M$ is itself.

Given a non-atomic macro $M$, we can find an atomic sub-macro $M'$ of $M$.
For example, from Figure~\ref{fig:trees-b}, we first find sub-macro $M' = M_1$.
We detach $M'$ from $M$ and define a macro rule:
\begin{align}\label{eqn:sub-macro-rule}
  c'_1[z_1] + \dots + c'_k[z_k] \to c'_\mathrm{out}[f(z_1,\dots,z_k)],
\end{align}
where $c'_1,\dots,c'_k$ are categories of the leaf nodes of $M'$,
and $f$ substitutes $z_1,\dots,z_k$ into the sub-macro $M'$.
The category $c'_\mathrm{out}$ is computed by serializing $M'$ as a string;
this way, if the sub-macro $M'$ appears in a different macro,
the category name will be shared.
Next, we substitute the subgraph $M'$ in $M$ by a placeholder node with name $c'_\mathrm{out}$.
The procedure is repeated on the new graph until the remaining macro is atomic. Finally, we define a single rule for the atomic macro.
The macro grammar uses the decomposed macro rules in replacement of
Rule~\eqref{eqn:atomic-macro-rule}.

For example, the macro in Figure~\ref{fig:trees-b} is decomposed into three macro rules:
\begin{align*}
Ent[z_1] &\to M_1[z_1],\\
Rel[z_1] + M_1[z_2] &\to M_2[\reverse{z_1}.\reverse{\tt Next}.z_1.z_2],\\
M_2[z_1] &\to Root[z_1].
\end{align*}
These correspond to the three atomic sub-macros $M_1$, $M_2$ and $M_3$ in
Figure~\ref{fig:trees-c}. The first and the second macro rules can be reused by
other macros.

Having defined macro rules,
we now describe how Algorithm~\ref{alg:learn-macro-grammar}
uses and updates the macro grammar
when processing each training example.

\subsection{Triggering macro rules}
\label{sec:trigger-macro-rules}

Throughout training, we keep track of a set $\mathcal{S}$ of training utterances that
have been associated with a consistent logical form. (The set $\mathcal{S}$ is
updated by Step~\ref{step:associate} of
Algorithm~\ref{alg:learn-macro-grammar}.) Then, given a training utterance $x$,
we compute its $K$-nearest neighbor utterances in $\mathcal{S}$, and select
all macro rules that were extracted from their associated logical forms. These macro rules are used to parse utterance $x$.

We use token-level Levenshtein distance as the distance metric for computing
nearest neighbors. More precisely, every utterance is written as a sequence of
lemmatized tokens $x = (x^{(1)},\dots,x^{(m)})$. After removing all determiners
and infrequent nouns that appear in less than 2\% of the
training utterances, the distance between two utterances $x$ and $x'$ is defined as the Levenshtein distance
between the two sequences. When computing the distance, we treat each word token as an atomic element. For example, the distance between ``highest score'' and ``best score'' is~1. Despite its simplicity, the Levenshtein distance
does a good job in capturing the structural similarity between utterances.
Table~\ref{table:nearest_neighbor_examples} shows that nearest neighbor
utterances often map to consistent logical forms with the same macro.

In order to compute the nearest neighbors efficiently, we pre-compute a sorted
list of $K_\text{max} = 100$ nearest neighbors for every utterance before
training starts. During training, calculating the intersection of this sorted
list with the set $\mathcal{S}$ gives the nearest neighbors required. For our
experiments, the preprocessing time is negligible compared to the overall
training time (less than 3\%), but if computing nearest neighbors is expensive, then
parallelization or approximate algorithms \cite[e.g.,][]{indyk2004approximate}
could be used.

\subsection{Updating model parameters}
\label{sec:train-and-predict}

Having computed the triggered macro rules $\mathcal{R}$, we combine them with the terminal rules $\mathcal{T}$ from the base grammar (e.g., for building
$Ent$ and $Rel$) to create a per-example grammar $\mathcal{R} \cup \mathcal{T}$ for the utterance $x$.
We use this grammar to generate logical forms using standard beam search. We
follow Section~\ref{sec:beam-search-and-learning} to generate a set
of candidate logical forms $\mathcal{Z}$ and update model
parameters.

However, we deviate from Section~\ref{sec:beam-search-and-learning} in one way.
Given a set $\mathcal{Z}$ of candidate logical forms for some training example $(x_i,w_i,y_i)$, we pick the
logical form $z_i^+$ with the highest model probability among consistent
logical forms, and pick $z_i^-$ with the highest model probability among
inconsistent logical forms, then perform a gradient update on the objective function:
\begin{gather}
  J(\theta) = \frac{1}{n}\sum_{i=1}^n \left( \log\frac{p_i^+}{p_i^-} \right) - \lambda \lones{\theta}, \label{eqn:new-objective-function} \\
  \begin{align*}
    \text{where\quad}  p_i^+ &= p_\theta(z_i^+ \mid x_i, w_i) & \\
    p_i^- &= p_\theta(z_i^- \mid x_i, w_i). &
  \end{align*}
\end{gather}
Compared to \eqref{eqn:objective-function}, this objective function only
considers the top consistent and inconsistent logical forms
for each example instead of all candidate logical forms. Empirically, we found
that optimizing \eqref{eqn:new-objective-function} gives a 2\% gain in 
prediction accuracy compared to optimizing \eqref{eqn:objective-function}.

\begin{table}
\centering
\scalebox{0.9}{
\begin{tabular}{|p{0.5\textwidth}|}
\hline
\emph{{\bf Who} ranked {\bf right after} Turkey?}\\
\emph{{\bf Who} took office {\bf right after} Uriah Forrest?}\\\hline
\emph{{\bf How many more} passengers flew to Los Angeles} \\
\emph{\quad {\bf than} to Saskatoon {\bf in} 2013?}\\
\emph{{\bf How many more} Hungarians live in the Serbian} \\
\emph{\quad Banat region {\bf than} Romanians {\bf in} 1910?}\\\hline
\emph{{\bf Which is} deeper{\bf ,} Lake Tuz {\bf or} Lake Palas Tuzla?}\\
\emph{{\bf Which} peak {\bf is} higher{\bf,} Mont Blanc {\bf or} Monte Rosa?}\\\hline
\end{tabular}
}
\caption{Examples of nearest neighbor utterances in the \wikitables\ dataset.}
\label{table:nearest_neighbor_examples}
\end{table}

\subsection{Updating the macro grammar}
\label{sec:update-macro-grammar}

If the triggered macro rules fail to find a consistent logical form, we fall
back to performing a beam search on the base grammar.
For efficiency, we stop the search either when a consistent logical
form is found, or when the total number of generated logical forms exceeds a
threshold~$T$.
The two stopping criteria prevent the search algorithm from spending too much
time on a complex example.
We might miss consistent logical forms on such examples,
but because the base grammar is only used for generating macro rules,
not for updating model parameters,
we might be able to induce the same macro rules from other examples.
For instance, if an example has an uttereance phrase that
matches too many knowledge base entries,
it would be more efficient to skip the example;
the macro that would have been extracted from this example
can be extracted from less ambiguous examples with the same question type.
Such omissions are not completely disastrous,
and can speed up training significantly.

When the algorithm succeeds in finding a consistent logical form $z$ using the base grammar, we derive
its macro $M$
following Section~\ref{sec:derive-macro}, then construct macro rules
following Section~\ref{sec:decompose-macro-rules}. These macro rules are added to the
macro grammar. We also associate the utterance $x$ with the consistent logical
form $z$, so that the macro rules that generate $z$ can be triggered by other
examples.
Parameters of the semantic parser are not updated in this case.

\subsection{Prediction}

At test time, we follow Steps~\ref{step:trigger}--\ref{step:generate} of Algorithm~\ref{alg:learn-macro-grammar} to generate a set $\mathcal{Z}$ of
candidate logical forms from the triggered macro rules, and then output the
highest-scoring logical form in $\mathcal{Z}$.
Since the base grammar is never used at test time, prediction is generally faster than training.

\section{Experiments}
\label{sec:experiments}

We report experiments on the \wikitables\ dataset~\cite{pasupat2015compositional}.
Our algorithm is compared with the parser trained only with the base grammar, the floating parser of~\citet{pasupat2015compositional}
(PL15), the Neural Programmer parser~\cite{neelakantan2016neural} and the
Neural Multi-Step Reasoning parser~\cite{haug2017neural}. Our algorithm not
only outperforms the others, but also achieves an order-of-magnitude speedup
over the parser trained with the base grammar and the parser in PL15.

\subsection{Setup}

\begin{table}
\centering
\scalebox{0.95}{
\begin{tabular}{|l|}
\hline
``\emph{Which driver appears the most?}''\\
{\small $\tt argmax(\reverse{Driver}.Type.Row,\reverse{\lambda x{\tt.count(Driver}.x)})$}
\\\hline
``\emph{What language was spoken more during}\\
\qquad\qquad \emph{the Olympic oath, English or French?}''\\
{\small $\tt argmax(English\sqcup French,\reverse{\lambda x{\tt.count(Language}.x)})$}\\\hline
``\emph{Who is taller, Rose or Tim?}''\\
{\small$\tt argmax(Rose\sqcup Tim,\reverse{\lambda x.{\tt\reverse{Num}.\reverse{Height}.Name}.x)})$}\\\hline
\end{tabular}
}
\caption{Several example logical forms our grammar
can generate that are not covered by PL15.}
\label{table:example-logical-form}
\end{table}

The dataset contains 22,033 complex questions on 2,108 Wikipedia tables. Each
question comes with a table, and the tables during evaluation are disjoint
from the ones during training. The training and test sets contain 14,152 and
4,344 examples respectively.\footnote{The remaining 3,537 examples were not
included in the original data split.} Following PL15, the development
accuracy is averaged over the first three 80-20 training data splits given in
the dataset package. The test accuracy is reported on the train-test data
split.

We use the same features and logical form pruning strategies as PL15,
but generalize
their base grammar.
To control the search space,
the actual system in PL15 restricts
the superlative operators \ttt{argmax} and \ttt{argmin} to be applied only on
the set of table rows. We allow these operators to be applied on the set of
tables cells as well, so that the grammar captures certain logical forms that
are not covered by PL15 (see Table~\ref{table:example-logical-form}).
Additionally, for terminal rule (\ref{eqn:terminal-anchored}), we allow
$f(span)$ to produce entities that approximately match the token span in addition to
exact matches. For example, the phrase ``Greenville'' can trigger both entities
\ttt{Greenville\_Ohio} and \ttt{Greensville}.

We chose hyperparameters using the first train-dev split. The beam size $B$ of beam
search is chosen to be $B=100$. The $K$-nearest neighbor parameter is chosen as
$K=40$. Like PL15, our algorithm takes 3 passes over the dataset for training. The maximum
number of logical forms generated in step~\ref{step:apply-base-grammar} of
Algorithm~\ref{alg:learn-macro-grammar} is set to $T=5{,}000$ for the first
pass. For subsequent passes, we set $T=0$ (i.e.,~never fall back to the base grammar)
so that we stop augmenting the macro grammar. During the first pass,
Algorithm~\ref{alg:learn-macro-grammar} falls back to the base grammar on roughly 30\% of the
training examples.

For training the baseline parser that only relies on the base grammar, we use the same beam size $B=100$, and take 3 passes over the dataset for training. There is no maximum constraint on the number of logical forms that can be generated for each example.

\subsection{Coverage of the macro grammar}

With the base grammar, our parser generates 13,700 partial logical forms on average for
each training example, and hits consistent logical forms on 81.0\% of the training examples.
With the macro rules from holistic triggering, these numbers become 1,300 and
75.6\%. The macro rules generate much fewer partial logical forms, but at the cost of slightly lower coverage.

However, these coverage numbers are computed based on finding any logical form that executes to the correct denotation.
This includes spurious logical forms,
which do not reflect the semantics of the question but are coincidentally consistent with the correct
denotation.
(For example,
the question ``\emph{Who got the same number of silvers as France?}'' on Table~\ref{table:wikitables-example}
might be spuriously parsed as $\tt \reverse{Nation}.\reverse{Next}.Nation.France$,
which represents the nation listed after France.)
To evaluate the ``true'' coverage, we sample 300 training examples and manually
label their logical forms. We find that on 48.7\% of these examples, the top
consistent logical form produced by the base grammar is semantically correct.
For the macro grammar, this ratio is also 48.7\%, meaning that the macro
grammar's effective coverage is as good as the base grammar.

The macro grammar extracts 123 macros in total. Among the 75.6\% examples that were covered by the macro grammar, the top 34 macros
cover 90\% of consistent logical forms. By examining the top 34 macros, we
discover explicit semantic meanings for 29 of them, which are described in detail in the supplementary material.

\subsection{Accuracy and speedup}

\begin{table}[t]
\begin{tabular}{|l|c|c|}
\hline
&Dev	&Test\\\hline
\citet{pasupat2015compositional}	&37.0\%	&37.1\%\\
\citet{neelakantan2016neural}	&37.5\%&	37.7\%\\
\citet{haug2017neural}	&-&	38.7\%\\\hline
This paper: base grammar	& {\bf 40.6\%}	&42.7\%\\
This paper: macro grammar	& 40.4\%	& {\bf 43.7\%} \\\hline
\end{tabular}
\caption{Results on \wikitables.}\label{table:compare-accuracies}
\end{table}

\newcommand{\midhead}[1]{\multicolumn{1}{|c|}{#1}}
\begin{table}[t]
\centering
\scalebox{0.9}{
\begin{tabular}{|l|r|r|r|}
\hline
& & \multicolumn{2}{|c|}{Time (ms/ex)} \\\cline{3-4}
& \midhead{Acc.} & \midhead{Train} & \midhead{Pred} \\\hline
PL15 & 37.0\%	& 619 & 645 \\\hline
Ours: base grammar & 40.6\%	& 1,117 & 1,150 \\\hline
Ours: macro grammar & 40.4\%  & 99 & 70 \\
~~no holistic triggering & 40.1\% &  361 & 369 \\
~~no macro decomposition & 40.3\% & 177 & 159 \\\hline
\end{tabular}}
\caption{Comparison and ablation study: the columns report averaged prediction accuracy, training time, and prediction time (milliseconds per example) on the three train-dev
  splits.}\label{table:compare-time}
\end{table}

\begin{figure*}[t]
\begin{subfigure}[b]{0.333\textwidth}\centering
\includegraphics[width=1.0\linewidth]{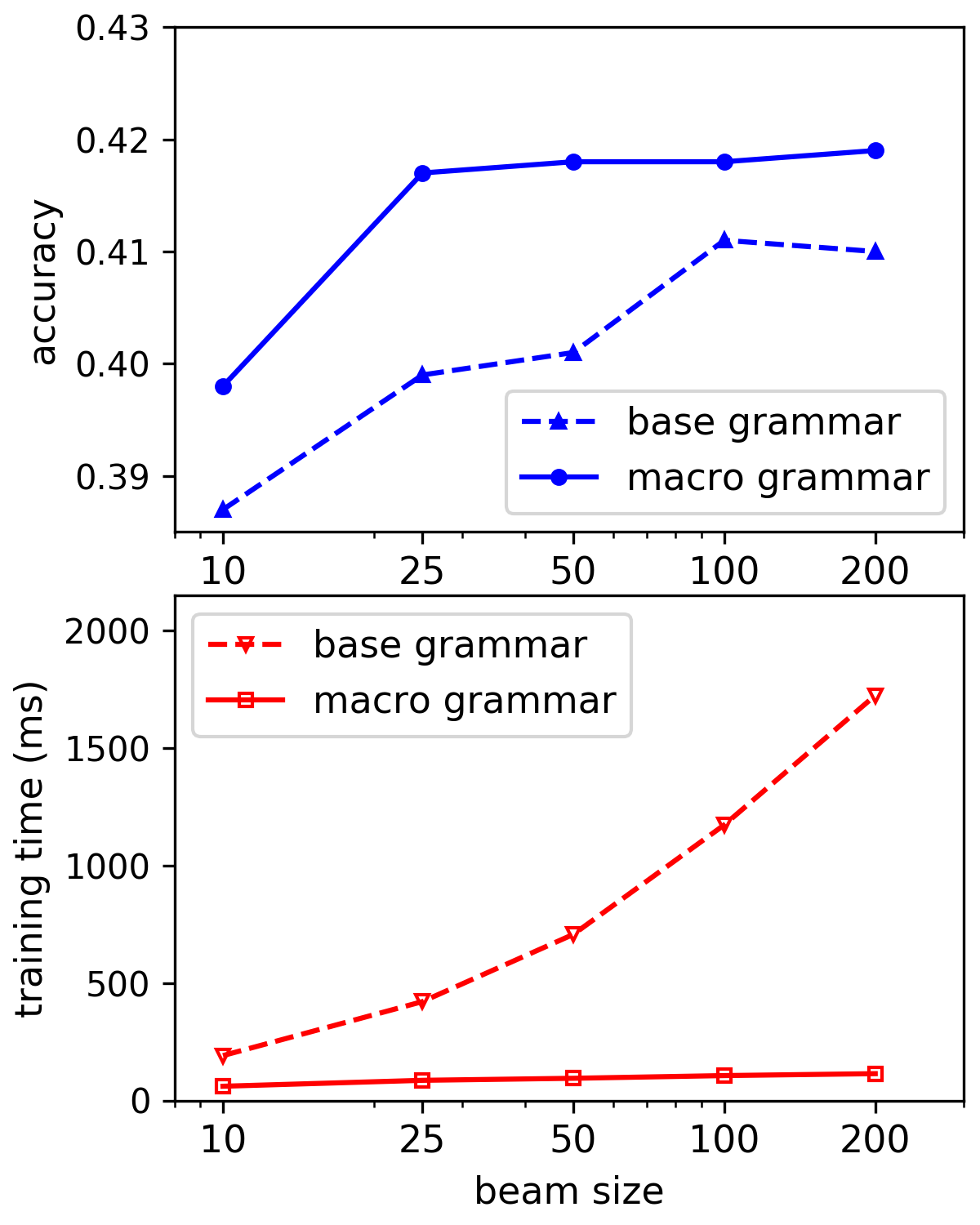}
\caption{Varying beam size}
\label{fig:vary-hyperparameters-a}
\end{subfigure}%
\begin{subfigure}[b]{0.333\textwidth}\centering
\includegraphics[width=1.0\linewidth]{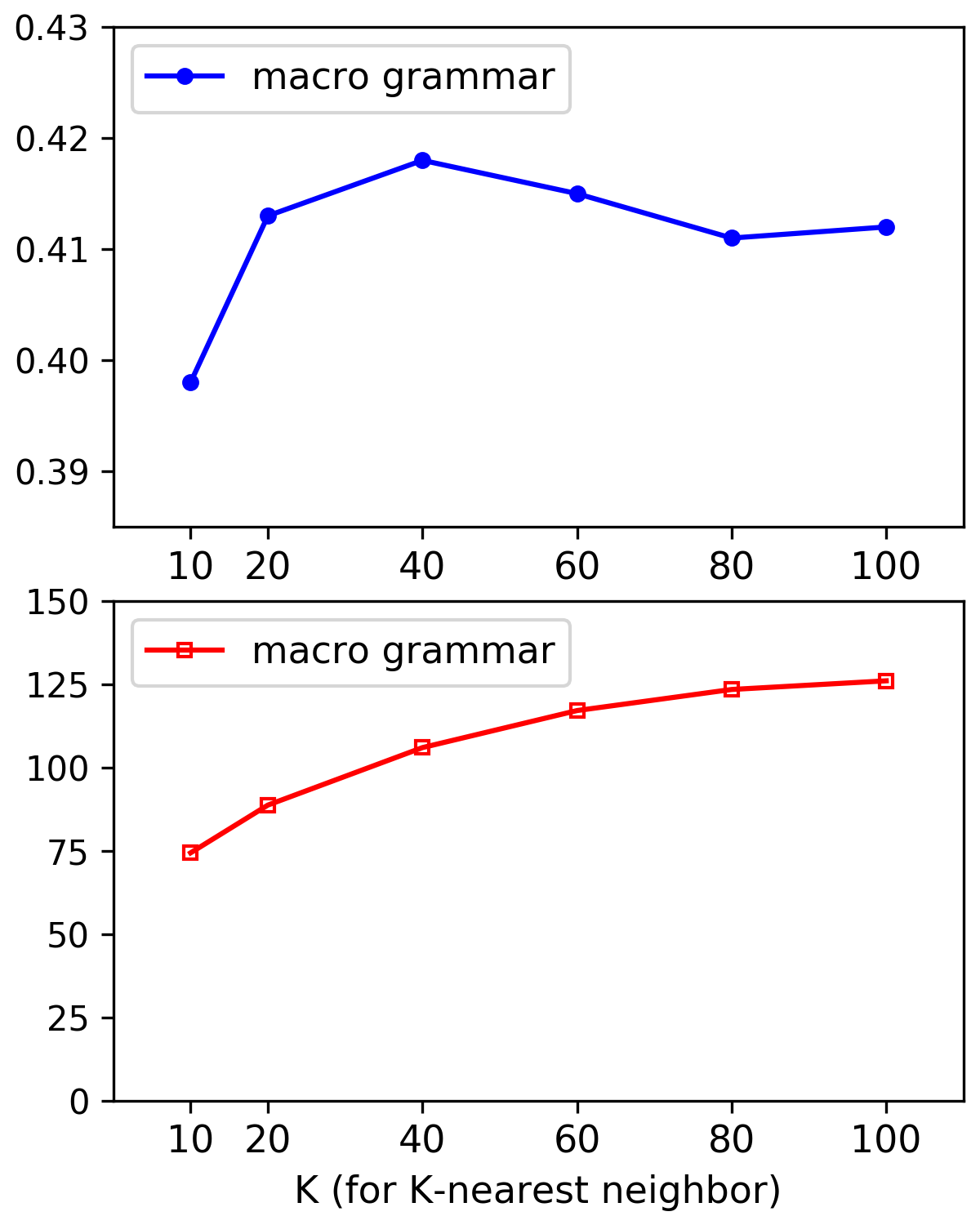}
\caption{Varying neighbor size}
\label{fig:vary-hyperparameters-b}
\end{subfigure}%
\begin{subfigure}[b]{0.333\textwidth}\centering
\includegraphics[width=1.0\linewidth]{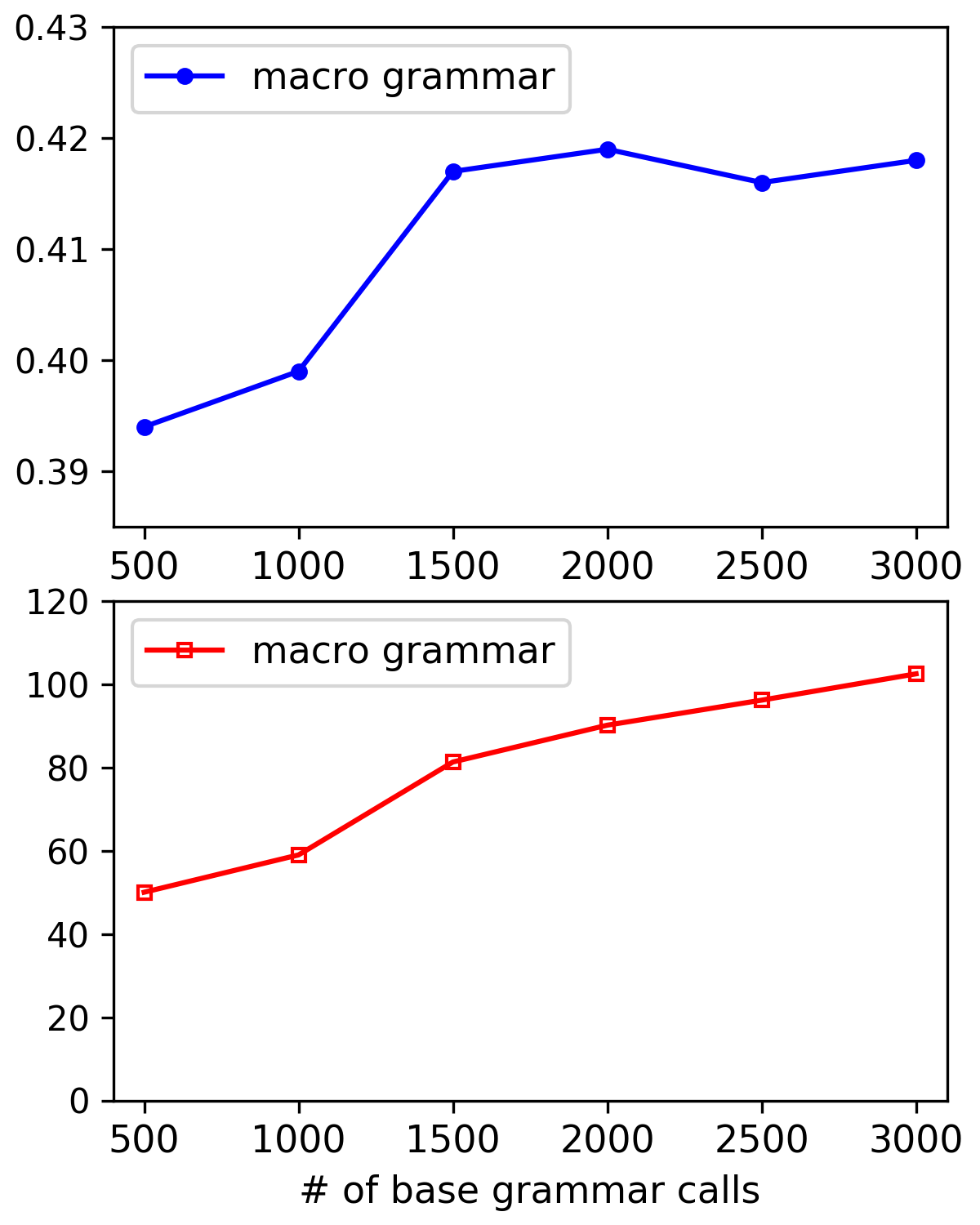}
\caption{Varying base grammar usage count}
\label{fig:vary-hyperparameters-c}
\end{subfigure}
\caption{Prediction accuracy and training time (per example) with various
  hyperparameter choices, reported on the first train-dev
  split.}\label{fig:vary-hyperparameters}
\end{figure*}

We report prediction accuracies in Table~\ref{table:compare-accuracies}. With a
more general base grammar (additional superlatives and approximate matching), and by optimizing the objective
function~\eqref{eqn:new-objective-function}, our base parser outperforms PL15 (42.7\% vs 37.1\%). Learning a macro grammar slightly improves the accuracy to 43.7\% on the test set.
On the three train-dev splits, the
averaged accuracy achieved by the base grammar and the macro grammar are close
(40.6\% vs 40.4\%).

In Table~\ref{table:compare-time}, we compare the training and prediction time
of PL15 as well as our parsers. For a fair comparison, we trained all parsers using
the SEMPRE toolkit~\cite{berant2013freebase} on a machine with Xeon 2.6GHz CPU
and 128GB memory without parallelization. The time for constructing the macro grammar is included as
part of the training time. Table~\ref{table:compare-time} shows that our parser
with the base grammar is more expensive to train than PL15. However, training
with the macro grammar is substantially more efficient than training with only the base grammar---
it achieves 11x speedup for training and 16x speedup for test time prediction.

We run two ablations of our algorithm to evaluate the utility of holistic triggering and macro decomposition.
The first ablation triggers all macro rules for parsing every utterance
without holistic triggering, while the
second ablation constructs Rule~\eqref{eqn:atomic-macro-rule} for every macro without decomposing it into smaller rules.
Table~\ref{table:compare-time} shows that both variants result in decreased efficiency.
This is because holistic triggering effectively prunes irrelevant macro
rules, while macro decomposition is important for efficient beam search and featurization.

\subsection{Influence of hyperparameters}

Figure~\ref{fig:vary-hyperparameters-a} shows that for all beam sizes,
training with the macro grammar is more efficient than training with the base
grammar, and the speedup rate grows with the beam size. The test time
accuracy
of the macro grammar is robust to varying beam sizes as long as
$B\geq 25$.

Figure~\ref{fig:vary-hyperparameters-b} shows the influence of the neighbor
size $K$. A smaller neighborhood triggers fewer macro rules, leading to
faster computation.
The accuracy peaks at $K=40$ then
decreases slightly for large $K$. We conjecture that the smaller number of
neighbors acts as a regularizer.

Figure~\ref{fig:vary-hyperparameters-c} reports an experiment where we limit
the number of fallback calls to the base grammar to $m$.
After the limit is reached, subsequent training examples that require fallback calls are simply skipped.
This limit means that the macro grammar will get augmented at most $m$ times during training.
We find
that for small $m$, the prediction accuracy grows with $m$, implying
that building a richer macro grammar improves the accuracy. For larger $m$,
however, the accuracies hardly change. According to the plot, a
competitive macro grammar can be built by calling the base grammar on less than
15\% of the training data.

Based on Figure~\ref{fig:vary-hyperparameters}, we can trade accuracy for
speed by choosing smaller values of $(B,K,m)$. With $B=50$, $K=40$ and
$m=2000$, the macro grammar achieves a slightly lower averaged development
accuracy ($40.2\%$ rather than $40.4\%$), but with an increased speedup of 15x (versus 11x) for training and 20x
(versus 16x) for prediction.

\section{Related work and discussion}
\label{sec:related-work}

A traditional semantic parser maps natural language phrases into partial logical forms
and composes these partial logical forms into complete logical forms.
Parsers define
composition based on a grammar formalism such as Combinatory Categorial Grammar (CCG)
\cite{zettlemoyer07relaxed,kwiatkowski11lex,kwiatkowski2013scaling,kushman2013regex,krishnamurthy2013jointly},
Synchronous CFG \cite{wong07synchronous}, and CFG
\cite{kate06krisp,chen11navigate,berant2013freebase,desai2016program}, while
others use the syntactic structure of the utterance to guide composition
\cite{poon09semantic,reddy2016transforming}. Recent neural semantic parsers
allow any sequence of logical tokens to be generated
\cite{dong2016logical,jia2016recombination,kocisk2016semantic,neelakantan2016neural,liang2017nsm,guu2017bridging}.
The flexibility of these composition methods allows arbitrary logical forms to be
generated, but at the cost of a vastly increased search space.

Whether we have annotated logical forms or not has dramatic implications
on what type of approach will work.  When logical forms are available,
one can perform grammar induction to mine grammar rules \emph{without search}
\citep{kwiatkowski10ccg}.  When only annotated denotations are available,
as in our setting, one must use a base grammar to define the output space of
logical forms.  Usually these base grammars come with many restrictions
to guard against combinatorial explosion \citep{pasupat2015compositional}.

Previous work on higher-order unification for lexicon induction
\cite{kwiatkowski10ccg} using factored lexicons \citep{kwiatkowski11lex} also learns logical form macros
with an online algorithm.
The result is a lexicon where each entry contains
a logical form template and a set of possible phrases for triggering the template.
In contrast, we have avoided binding grammar rules to particular phrases
in order to handle lexical variations.
Instead, we use a more flexible mechanism---holistic triggering---to determine
which rules to fire.  This allows us to generate logical forms for utterances
containing unseen lexical paraphrases or where the triggering is spread throughout the sentence.
For example, the question ``\emph{Who is X, John or Y}''
can still trigger the correct macro extracted from the last example in Table~\ref{table:example-logical-form}
even when \emph{X} and \emph{Y} are unknown words.

Our macro grammars bears some resemblance to
adaptor grammars \cite{johnson06adaptor} and fragment grammars \cite{odonnell11fragment},
which are also based on the idea of caching useful chunks of outputs.
These generative approaches aim to solve the modeling problem
of assigning higher probability mass to outputs that use reoccurring parts.
In contrast, our learning algorithm uses caching as a way to constrain the search space for computational efficiency;
the probabilities of the candidate outputs are assigned by a separate discriminative model.
That said, the use of macro grammars does have a small positive modeling contribution,
as it increases test accuracy from 42.7\% to 43.7\%.

An orthogonal approach for improving search efficiency is to adaptively choose
which part of the search space to explore. For example,
\newcite{berant2015agenda} uses imitation learning to strategically search for
logical forms.
Our holistic triggering method, which selects macro rules based on the
similarity of input utterances, is related to the use of paraphrases
\cite{berant2014paraphrasing,fader2013paraphrase} or string kernels
\cite{kate06krisp} to train semantic parsers.
While the input similarity measure is critical for scoring logical forms in these previous works, we use
the measure only to retrieve candidate rules, while scoring is done by a separate
model.  The retrieval bar means that our similarity metric can be quite crude.

\section{Summary}

We have presented a method for speeding up semantic parsing via macro grammars.
The main source of efficiency is the decreased size of the logical form space.
By performing beam search on a few macro rules associated with the $K$-nearest neighbor utterances via holistic triggering,
we have restricted the search space to semantically relevant logical forms.
At the same time, we still maintain coverage over the base logical form space
by occasionally falling back to the base grammar
and using the consistent logical forms found to enrich the macro grammar.
The higher efficiency allows us expand the base grammar
without having to worry much about speed:
our model achieves a state-of-the-art accuracy
while also enjoying an order magnitude speedup.

\paragraph{Acknowledgements.}
We gratefully acknowledge Tencent for their
support on this project.

\paragraph{Reproducibility.} Code, data, and experiments for this paper are available on CodaLab platform:
\href{https://worksheets.codalab.org/worksheets/0x4d6dbfc5ec7f44a6a4da4ca2a9334d6e/}{
\small{\tt https://worksheets.codalab.org/worksheets/} {\tt 0x4d6dbfc5ec7f44a6a4da4ca2a9334d6e/}}.
\normalsize

\bibliographystyle{emnlp_natbib}
\bibliography{refdb/all.bib}

\newcommand{\template}[3]{{\bf Macro}: {{#1}}\\{\bf Description}: {#2}.\\{\bf Example}: \emph{{#3}}}

\onecolumn
\section*{Supplementary material: macro analysis}

The macro grammar extracts 123 macros from the WikiTableQuestions dataset, covering consistent logical forms for 75.6\% examples. Let the frequency of a macro be defined as the number of highest-scoring consistent logical forms that it generates. We plot the frequency of all macros, sorted in decreasing order:\\

\begin{center}
\includegraphics[width=0.9\textwidth]{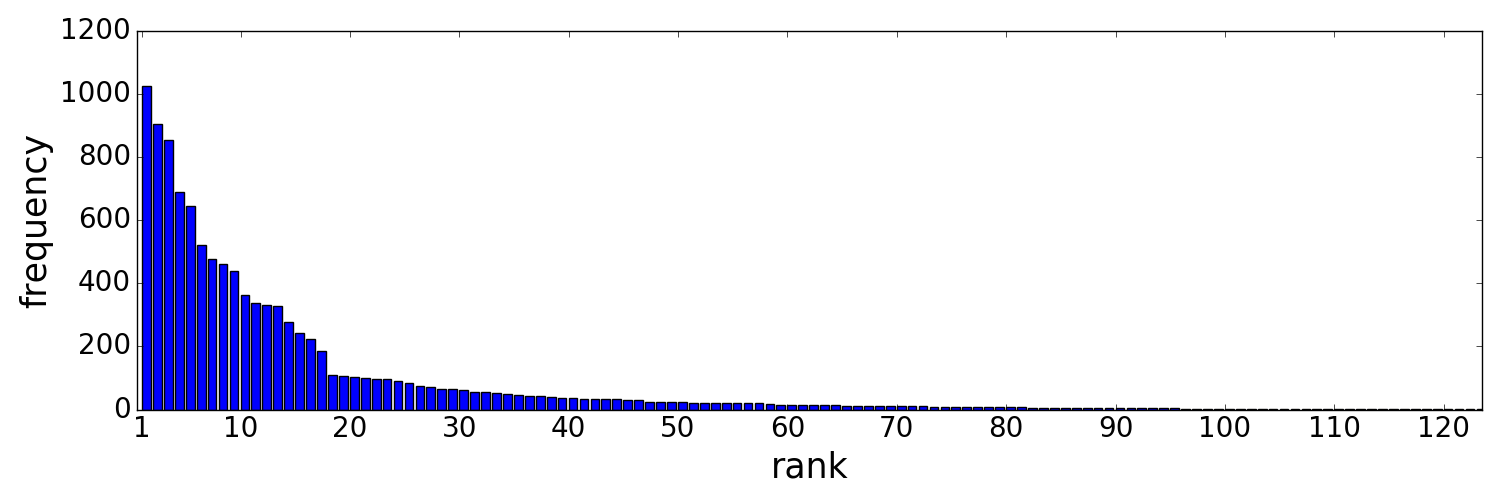}
\end{center}

As demonstrated by the plot, the top 20 macros cover 80\% total frequency, and the top 34 macros cover 90\% total frequency. It suggests that a small fraction of macros capture most examples' consistent logical forms. By manually examining the top 34 macros, we find that 29 of them have explicit semantics. These macros correspond to abstract operations on the table, but when their slots are filled with concrete entities and relations\footnote{A macro could have four categories of slots: \vspace*{-.8em}
\begin{itemize}
  \item \{Col\#x\} represents a column relation: {\tt Name}, {\tt Rank}, {\tt Venue}, etc. \vspace*{-.8em}
  \item \{Prop\#x\} represents a property relation: {\tt Number}, {\tt Year}, {\tt Date}, etc. \vspace*{-.8em}
  \item \{Compare\#x\} represents a comparative relation: {\tt >}, {\tt <}, {\tt >=}, {\tt <=}. \vspace*{-.8em}
  \item \{Ent\#x\} represents an entity: {\tt Turkey}, {\tt (number 2)}, {\tt (year 1998)}, etc. \vspace*{-.8em}
\end{itemize}}, they can be phrased in meaningful natural language utterances. Below, we interpret the meaning of each macro using examples from the WikiTableQuestions dataset:

\begin{enumerate}
\item \template{{\bf count}(\{Col\#1\}.\{Ent\#2\})}
{the number of rows whose column \{Col\#1\} matches \{Ent\#2\}}
{how many records were set in Beijing ?}

\item \template{{\bf R}[\{Col\#1\}].\{Col\#2\}.\{Ent\#3\}}
{select rows whose column \{Col\#2\} matches \{Ent\#3\}, then return all entities in column \{Col\#1\}}
{what mine is in the town of Timmins?}

\item \template{{\bf R}[\{Prop\#1\}].{\bf R}[\{Col\#2\}].\{Col\#3\}.\{Ent\#4\}}
{select rows whose column \{Col\#3\} matches \{Ent\#4\}, then return property \{Prop\#1\} for all entities in column \{Col\#2\}}
{what is the number of inhabitants living in Predeal?}

\item \template{{\bf count}(\{Col\#1\}.\{Prop\#2\}.\{Compare\#3\}.\{Ent\#4\})}
{the number of rows satisfying some comparative constraint}
{how many directors served more than 3 years?}

\item \template{{\bf R}[\{Col\#1\}].{\bf argmax}(Type.Row, {\bf R}[$\lambda x$.{\bf R}[\{Prop\#2\}].{\bf R}[\{Col\#3\}].$x$])}
{select the largest value in column \{Col\#3\}, then for the associated row, return entities in column \{Col\#1\}}
{which team scored the most goal?}

\item \template{{\bf R}[\{Col\#1\}].{\bf R}[Next].\{Col\#1\}.\{Ent\#2\}}
{return the entity right below \{Ent\#2\}}
{who ranked right after Turkey?}

\item \template{{\bf R}[\{Col\#1\}].{\bf argmin}(Type.Row, {\bf R}[$\lambda x$.{\bf R}[\{Prop\#2\}].{\bf R}[Col\#3].$x$])}
{select the smallest value in column \{Col\#3\}, then for the associated row, return entities in column \{Col\#1\}}
{which team scored the least goal?}

\item \template{{\bf R}[\{Col\#1\}].{\bf argmin}(Type.Row, index)}
{return column \{Col\#1\} of the first row}
{which president is listed at the top of the chart ?}

\item \template{{\bf count}(\{Col\#1\}.{\bf argmax}({\bf R}[\{Col\#1\}].Type.Row, {\bf R}[$\lambda x$.{\bf count}(\{Col\#1\}.$x$)]))}
{N/A}
{\rm N/A}

\item \template{{\bf count}(\{Col\#1\}.{\bf argmin}({\bf R}[\{Col\#1\}].Type.Row, {\bf R}[$\lambda x$.{\bf count}(\{Col\#1\}.$x$)]))}
{N/A}
{\rm N/A}

\item \template{{\bf R}[\{Col\#1\}].{\bf argmax}(Type.Row, index)}
{return column \{Col\#1\} of the last row}
{which president is listed at the bottom of the chart ?}

\item \label{item:next-entity}\template{{\bf R}[\{Col\#1\}].Next.{\bf argmin}({\bf R}[\{Col\#1\}].\{Ent\#2\}, index)}
{return the entity right above \{Ent\#2\}}
{who is listed before Jon Taylor?}

\item \template{{\bf count}(Type.Row)}
{the total number of rows}
{what is the total number of teams?}

\item \template{{\bf argmax}({\bf R}[\{Col\#1\}].Type.Row, {\bf R}[$\lambda x$.{\bf count}(\{Col\#1\}.$x$)]))}
{return the most frequent entity in column \{Col\#1\}}
{which county has the most number of representatives?}

\item \template{{\bf sub}({\bf R}[\{Prop\#1\}].{\bf R}[\{Col\#2\}].\{Col\#3\}.\{Ent\#4\}, {\bf R}[\{Prop\#1\}].{\bf R}[\{Col\#2\}]\\.\{Col\#3\}.\{Ent\#5\})}
{Given two entities, calculate the difference for some property}
{how many more passengers flew to Los Angeles than to Saskatoon?}

\item \template{{\bf argmax}({\bf or}(\{Ent\#1\}, \{Ent\#2\}), {\bf R}[$\lambda x$.{\bf R}[\{Prop\#3\}].{\bf R}[\{Col\#4\}].\{Col\#5\}.$x$]))}
{among two entities, return the one that is greater in some property}
{which is deeper, Lake Tuz or Lake Palas Tuzla?}

\item \template{{\bf R}[\{Col\#1\}].{\bf argmin}(\{Col\#1\}.{\bf or}(\{Ent\#1\}, \{Ent\#2\}), index)}
{N/A}
{\rm N/A}

\item \template{{\bf R}[\{Col\#1\}].\{Col\#2\}.\{Prop\#3\}.\{Compare\#4\}.\{Ent\#4\}}
{select rows whose property satisfies a comparative constraint, then return all entities in column \{Col\#1\}}
{which artist have released at least 5 albums?}

\item \template{{\bf max}({\bf R}[\{Prop\#1\}].{\bf R}[\{Col\#2\}].Type.Row)}
{return the maximum value in column \{Col\#2\}}
{what is the top population on the chart?}

\item \template{{\bf R}[\{Prop\#1\}].{\bf R}[\{Col\#2\}].{\bf argmin}(Type.Row, index)}
{return a property in the first row's column \{Col\#2\}}
{what is the first year listed?}

\item \template{{\bf R}[\{Col\#1\}].{\bf argmin}(\{Col\#2\}.\{Prop\#3\}.\{Compare\#4\}.\{Ent\#5\}, index)}
{select the first row that satisfies a comparative constraint, then return its column \{Col\#1\}}
{what is the first creature after page 40?}

\item \template{{\bf R}[\{Col\#1\}].{\bf argmin}(\{Col\#2\}.\{Ent\#3\}, index)}
{select the first row whose column \{Col\#2\} matches entity \{Ent\#3\}, then return its column \{Col\#1\}}
{who is the top finisher from Poland?}

\item \template{{\bf R}[\{Col\#1\}].\{Col\#2\}.\{Prop\#3\}.\{Ent\#4\}}
{select rows whose column \{Col\#2\} matches some property, then return all entities in column \{Col\#1\}}
{who is the only one in 4th place?}

\item \template{{\bf R}[\{Prop\#1\}].{\bf R}[\{Col\#2\}].{\bf argmax}(Type.Row, index)}
{return a property of column \{Col\#2\} of the first row}
{what is the first year listed?}

\item \template{{\bf R}[\{Col\#1\}].Next.\{Col\#1\}.\{Ent\#2\}}
{same as macro~\ref{item:next-entity}}
{\rm same as macro~\ref{item:next-entity}}

\item \template{{\bf min}({\bf R}[\{Prop\#1\}].{\bf R}[\{Col\#2\}].Type.Row)}
{return the minimum value in column \{Col\#2\}}
{what is the least amount of laps completed?}

\item \template{{\bf R}[\{Col\#1\}].{\bf argmax}(\{Col\#2\}.\{Ent\#3\}, index)}
{select the last row whose column \{Col\#2\} matches entity \{Ent\#3\}, then return its column \{Col\#1\}}
{what was the last game created by Spicy Horse?}

\item \template{{\bf count}(\{Col\#1\}.{\bf or}(\{Ent\#2\}, \{Ent\#3\}))}
{the number of rows whose column \{Col\#1\} matches either \{Ent\#2\} or \{Ent\#3\}}
{how many total medals did switzerland and france win?}

\item \label{item:prop-argmax-prop}\template{{\bf R}[\{Prop\#1\}].{\bf R}[\{Col\#2\}].{\bf argmax}(Type.Row, {\bf R}[$\lambda x$.{\bf R}[\{Prop\#3\}].{\bf R}[\{Col\#4\}].$x$])}
{select the largest value in column \{Col\#4\}, then for the associated row, return a property of column \{Col\#2\}}
{what year had the highest unemployment rate?}

\item \template{{\bf count}(\{Col\#1\}.\{Prop\#2\}.\{Ent\#3\})}
{the number of rows whose column \{Col\#1\} matches a property \{Ent\#3\}}
{how many people were born in 1976?}

\item \template{{\bf count}({\bf argmin}(Type.Row, {\bf R}[$\lambda x$.{\bf R}[\{Prop\#1\}].{\bf R}[Col\#2].$x$]))}
{N/A}
{\rm N/A}

\item \template{{\bf sub}({\bf count}(\{Col\#1\}.\{Ent\#2\}), {\bf count}(\{Col\#1\}.\{Ent\#3\}))}
{Given two entities, calculate the difference of their frequencies in column \{Col\#1\}}
{how many more games were released in 2005 than 2003?}

\item \template{{\bf R}[\{Prop\#1\}].{\bf R}[\{Col\#2\}].{\bf argmax}(Type.Row, {\bf R}[$\lambda x$.{\bf R}[\{Prop\#1\}].{\bf R}[Col\#3].$x$])}
{same as macro~\ref{item:prop-argmax-prop}, but with an additional constraint that the two properties in the logical form must be equal}
{which game number has the most attendance?}

\item \template{{\bf R}[\{Col\#1\}].{\bf R}[Next].{\bf argmin}(Type.Row, index)}
{N/A}
{\rm N/A}
\end{enumerate}

\end{document}